\documentclass[manuscript,screen]{acmart}
\AtBeginDocument{%
  \providecommand\BibTeX{{%
    \normalfont B\kern-0.5em{\scshape i\kern-0.25em b}\kern-0.8em\TeX}}}

\setcopyright{acmcopyright}
\copyrightyear{2023}
\acmYear{2023}

\acmConference[SARs: TMI at CHI 2023]{}{April 28, 2023}{Hamburg, Germany}
\acmBooktitle{1st Workshop on Socially Assistive Robots as Decision Makers: 
Transparency, Motivations, and Intentions at CHI 2023, April 23--28, 2023, Hamburg, Germany} 
\acmPrice{15.00}
\acmISBN{978-1-4503-XXXX-X/18/06}

\begin{document}

\title{Addressing Potential Pitfalls of SAR Assistance on the Aging Population} 

\author{Emilyann Nault}
\email{en27@hw.ac.uk}
\orcid{0000-0003-3981-5795}
\author{Ronnie Smith}
\orcid{0000-0002-4014-5736}
\affiliation{%
  \institution{Heriot-Watt University \& University of Edinburgh}
  \city{Edinburgh}
  \country{United Kingdom}
}

\author{Lynne Baillie}
\orcid{0000-0002-2514-5981}
\email{l.baillie@hw.ac.uk}
\affiliation{%
  \institution{Heriot-Watt University}
  \city{Edinburgh}
  \country{United Kingdom}
}

\renewcommand{\shortauthors}{Nault et al.}

\begin{abstract}
In the field of Human Robot Interaction (HRI), socially assistive robots are being investigated to see if they can help combat challenges that can come with aging by providing different forms of support to older adults. As a result, it is imperative that the HRI community are aware of the potential pitfalls that can occur such as over-attachment, over-reliance, and increased isolation. This position paper argues designers should (a) avoid pitfalls that can lead to a negative impact on decline, and (b) leverage SAR decision making to avoid the pitfalls while attaining the benefits of this technology. Finally, we describe the concept for a framework as a starting point for addressing the concerns raised in this paper. 
\end{abstract}



\maketitle

\section{Introduction}

There has been substantial research in recent years towards the development of Socially Assistive Robots (SARs) to assist older adults (individuals aged 65 and over \cite{coghlan2021companion}). Older adults have specific considerations due to the nature of the physical and cognitive decline that comes along with aging. Researchers should be aware of pitfalls associated with the use of SARs, particularly when it comes to how the technology may impact the decline of their users, and should be equipped to take precautionary measures to avoid them. This position paper argues the HRI community should (a) avoid pitfalls that can lead to a negative impact on decline, and (b) leverage SAR decision making to avoid the pitfalls while attaining the benefits of this technology.

SARs can aid older adults in many areas including activities of daily living (ADL) \cite{rockwood_global_2005}, cognitive training support \cite{alnajjar2019cct, nault2022ro-man}, and providing companionship \cite{coghlan2021companion}. Oftentimes the goal of this area of research is to enable individuals to live independently for longer \cite{prescott_robotics_2017, coghlan2021companion}. All of these use cases require social interaction between the older adult and the robot. Section \ref{sar_decision_making} will address specific areas of concern regarding how SARs can negatively impact decline, including unhealthy attachment and reliance on the robot, and how SARs may result in increased isolation. It will further describe how SAR decision-making can be utilized to avoid these pitfalls while maximizing the benefits of this technology. Lastly, a concept for a framework to address these concerns will be proposed.
 
\section{Potential Pitfalls}
\label{sar_decision_making}


\subsection{Attachment}
\label{attachment}

Attachment plays an important role for those with dementia in providing feelings of security. A 30-day trial of doll therapy with older adults with dementia in a care home resulted in improvements in a neuropsychiatric evaluation, reduction in the distress measure, and better coping with separation from nurses compared to the control \cite{molteni2022doll}. SARs could offer further attachment through the provision of social engagement, particularly when they are being utilized to offer companionship. Robotic dogs have been shown to elicit similar degrees of attachment to living dogs for older adults \cite{banks2008dog_attachment}. However, as with any technology, it can break, and new versions are constantly being developed, sometimes resulting in older models no longer being supported. 

To avoid potentially negative consequences when the older adult is separated from the robot, we believe it falls to the robotics industry to improve the reliability of their products. This is easier said than done and will likely be a lengthy process, but it can be addressed in a variety of ways. Maintaining stable versions of software and validating with the target demographic prior to release can limit difficulties that arise from software updates. Also, designing physically robust robots (e.g., water-resistant, easy to clean) can reduce the need for repairs. Technical support can allow the older adult, carers, or family members to assist with repairs, or even provide services where the technician comes to the home. When all else fails, the older adult (or institution such as a care home) may not have the means to purchase another robot. This highlights the importance of providing robust, cost-effective solutions.

For someone with a high attachment to a SAR, removing the robot could be emotionally detrimental to the individual. This needs to be taken into consideration by the research community when running evaluations with SARs. Chen \textit{et al.} \cite{chen2020paro_attachment} found some older adults had difficulty when Paro, a robotic fur seal, was removed after 24 hours. For one participant, this negatively impacted their sleep for a few days after the trial. These participants did not have mild cognitive impairment or dementia. This begs the question of the impact on users when the SAR is taken away after long-term interactions, particularly with those experiencing age-related cognitive decline. 

If the user is overly attached to the SAR (e.g., to the point where it results in isolation), the first stage would be for the SAR to identify this unhealthy attachment. This could be from comparing the older adults' typical social engagement and encouraging through social interaction for the user to engage with others apart from the SAR. In severe cases, there should be a means for the SAR to notify appropriate services (e.g., family or doctor). While this has not been explicitly studied, recent literature has called for ethical means of managing this attachment in the research context before SARs are made more widely available \cite{ostrowski2022attachment}.

\subsection{Reliance}

Domestic robots are a special type of SAR which support individuals in their daily life through practical support around the home (fetching objects, opening doors, etc.) and software-only services (e.g. pill reminders). They are also becoming increasingly `proactive', i.e., making autonomous actions to assist the user at home \cite{grosinger_robots_2019}\cite{harman_action_2020}\cite{saunders_teach_2016}. By design, such robots may automate tasks ordinarily performed by the person they support. As such, it is important to consider whether these robots could negatively impact user health trajectories. While it is a common aim for SARs to physically and cognitively engage the user to reduce dependency and increase self-efficacy \cite{metzelthin_doing_2017}, robots which also automate routine tasks may inadvertently contradict this.

Individuals who are `mildly frail', based on the Clinical Frailty Scale, are at a stage of life where they are more evidently slowing down and require some help with instrumental ADLs (IADLs) \cite{rockwood_global_2005}. At this stage, individuals may struggle to achieve healthy levels of daily activity, e.g. coming in under the suggested 3,000 daily steps for those over 60 \cite{paluch_daily_2022}. Furthermore, IADLs like cooking require multiple areas of cognition (e.g., executive function, memory, and attention). A reduction in this engagement may negatively impact their cognitive decline as well. Therefore, it is especially important to consider the potential negative impact of individuals becoming dependent on domestic robots. We argue that this may have practical implications on their physical and cognitive health. In particular, over-reliance on the robot may lead to increased sedentariness and therefore negatively impact a host of mortality factors.

It may be possible to avoid over-reliance by ensuring the user has ultimate control over the robot and can personalize it according to their needs. Users could be made aware of exactly what their robot can do and could then specify which services they want to enable, rather than have the robot automatically try to perform them all \cite{saunders_teach_2016}. The downside is that the user may find it difficult to learn how to program the robot. It is also possible to personalize based on specific factors relevant to the individual. SARs already exist which take into account sociodemographic factors in their decision making \cite{flandorfer_population_2012}, and so future SARs could likewise take into account relevant health factors. For instance, for robots that support individuals at home, behavior may differ based on user frailty mobility. For some individuals, it may seek to collaborate and actively motivate the user to participate in activities. For those who have particular trouble with certain tasks, the robot will know to perform those tasks independently.

Conversely, there is little evidence to suggest negative impact of SARs on physical and cognitive rehabilitation. SARs can be useful as a rehabilitation aid, primarily in encouraging uptake of activities and to provide support throughout. For example, both \cite{swift-spong_effects_2015} and \cite{feingold-polak_robot_2021} feature SARs which coach the user on physical functional tasks. The SARs had a positive impact on the rehabilitation of the individuals, and in the latter, the SAR lead to greater motivation versus a computer-based system. Likewise, a SAR has been used in cardiac rehabilitation, where the robot provided encouragement and personalized feedback on progress in a positive way \cite{irfan_using_2020}. Compared to a control group using a tablet-based interface, the SAR was found to be more usable and trustworthy. Cognitive training has been shown to assist in maintaining (and sometimes improving) cognition in older adults \cite{alnajjar2019cct}. This therapy is oftentimes delivered via a computer (known as computerized cognitive training, or CCT). A 2019 systematic review \cite{alnajjar2019cct} found that while CCT is effective at improving cognitive function in older adults, administering this therapy via a robot improves their overall cognition. This is true for both older adults without age-related cognitive impairment as well as those with mild cognitive impairment. Overall, it would appear that in the worst cases a SAR offers no additional benefit over rehabilitation without a robot. However, work up to now has not directly addressed reliance on SARs for rehabilitation (i.e., long-term in-the-wild experiments).

\subsection{Isolation}

Social isolation and perceived isolation are associated with overall cognitive function and other areas of cognition including memory, attention, executive functioning, and language skills \cite{dinapoli2014isolation}. Older adults remain isolated for a variety of reasons, including the individual enjoying being solitary as well as some wanting company but fearing they will stay too long \cite{coghlan2021companion}. Increased isolation can also be caused by an unhealthy attachment to SARs, as discussed in Section \ref{attachment}. This can lead to the older adult over-attributing emotional capabilities to the robot \cite{mataric2016attachment}. For instance, the older adult may feel that the robot will miss them if they leave the house, resulting in them becoming more isolated.

One concern of older adults is the loss of independence, and consequently, that they will become dependent on others \cite{coghlan2021companion}. Interviews with older adults surrounding acceptance factors towards companion robots to combat loneliness and isolation found that some expressed concerns over others seeing them interacting with the robot  \cite{coghlan2021companion}. This was due to a fear of being perceived as declining or dependent on the robot. Particularly, participants thought it may be embarrassing, or even shameful, to interact with such a technology. For example, this could lead to the older adult not inviting people over to their residence.

Some older adults worry SARs may have unintended effects, i.e., ``I believe the use of the robot will restrict my autonomy'' \cite{pino_are_2015}. SARs can combat this by making decisions that actively promote user autonomy. For example, granting the older adult control over the SAR embodiment (e.g., a smaller robot could be stored away when people visit) and the length of the interaction could promote further social engagement while limiting isolation. Releasing some of the decision-making of the SAR into the hands of the user can grant them autonomy, allowing them to gain a greater sense of independence and thereby decreasing the feeling of reliance on the technology. Moreover, ethicists have argued that the integration of autonomy into robotic design can directly alleviate these concerns \cite{sharkey2012robot_ethics_autonomy}.

\section{Potential Framework}

We see that there is potential for a framework that will guide SAR designers in considering the potential pitfalls of the agents they create. By answering questions about the intended purpose of their SAR, the framework should prompt the designer to think about the stakeholders involved and what their relationship to the SAR will be (older adults using the SAR, formal and informal carers, etc.). Furthermore, it should enable the designer to consider the effect their SAR might have on a variety of factors, including those highlighted here but also on specific health factors.

For instance, part of the framework may involve asking questions of the designer against a scale, such as ``\textit{Is this a robot you would be happy for others to see you using?}'', ``\textit{Does the SAR fit practically into the user's environment and daily routine?}'', or ``\textit{Will using this robot involve the user undertaking less physical movement?}''. Depending on the answer, it is possible to offer relevant advice and feedback and to encourage thought on mitigation, e.g. if they would not be happy for people to see them using it, consider the impact it may have on the sense of isolation and autonomy of its user.

\section{Conclusion}

In this paper, we have highlighted that while SARs have great potential to benefit older adults, there are some pitfalls in relation to attachment, reliance, and isolation that can negatively impact their decline. We argue the HRI community should be mindful of and actively work to avoid these pitfalls to ensure SARs create only their intended positive impact. While further work is required to decipher how to manage unhealthy attachment to SARs, we propose monitoring and encouragement towards healthy behaviors are steps in the correct direction. For over-reliance, personalization of SARs (whether performed by the end-user or those who care for them) can ensure robots provide only the most appropriate assistance for each individual---i.e. a one size fits all approach is often not the ideal choice. Lastly, to combat increased isolation, we suggest designing SAR decision making to promote the autonomy of the older adult. While it is already understood that older adults seek control over their own assistive technology, we posit that focusing on individual autonomy also has positive implications for the effectiveness of SARs. We finally propose a concept for a framework as a starting point moving forward to achieve the goals of this paper: (a) avoid pitfalls that can lead to a negative impact on decline, and (b) leverage SAR decision making to avoid the pitfalls while attaining the benefits of this technology.







\begin{acks}
Emilyann Nault and Ronnie Smith are supported by the EPSRC Centre for Doctoral Training in Robotics and Autonomous Systems (EP/S023208/1 and EP/L016834/1, respectively). Ronnie Smith is also supported by EMERGENCE: Emergence of Healthcare Robots from Labs into Service (EP/W000741/1). Professor Lynne Baillie is supported by the Feather project (EP/W031493/1).
\end{acks}
  
\bibliographystyle{ACM-Reference-Format}
\bibliography{bibliography}


\end{document}